# نحو تعريب أفضل لبرينستون ووردنت

عبد الحكيم فريحات

جامعة ترينتو - الجمهورية الإيطالية

## ملخص

مع ازدياد أهمية برينستون ووردنت كمعجم دلالي في مجال معالجة اللغات الطبيعية تبرز الحاجة لتعريب هذا المعجم وضمان جودة التعريب. للآن نجد أن هناك جهدا متواضعا في تعريب هذا المعجم حيث أننا نجد أن محاولات التعريب حتى هذه اللحظة متواضعة كما وكيفا. بنفس الوقت لا نجد أي دراسات تهتم بصحة وجودة التعريب ووضعها بالسياق الثقافي الخاص باللغة العربية. نقترح في هذا البحث إطارا عاما لتعريب برينستون ووردنت ونشرح المراحل والخطوات التي نقترح اتباعها للوصول لتعريب عالي الجودة دون الاخلال بالسياق الثقافي للغة العربية. في نهاية هذا البحث سنقدم تجربتنا في استخدام الآلية المقترحة للتعريب والنتائج التي حصلنا عليها في تعريب عشرة آلاف سنست من برينستون وورد نت.

## 1  مقدمة:

برينستون ووردنت [1,2] هو معجم دلالي للغة الإنجليزية ينظم المفردات بناء على دلالاتها باستخدام علاقات لغوية ودلالية وليس بناء على الترتيب الأبجدي كما هو الحال في المعاجم اللغوية. تم تطوير هذا المعجم في جامعة برينستون كمشروع بحثي يهدف إلى تطوير قاموس يستند في تنظيم الكلمات والعلاقات الدلالية بينها على النظريات اللغوية النفسية وفهم كيفية تنظيم العقل البشري للمفردات والمعاني [3].

كان لووردنت دور مهم وفعال في تطوير وتحسين برامج معالجة اللغة الطبيعية [4] الخاصة باللغة الإنجليزية الأمر الذي جذب أنظار العلماء والباحثين اللغويين نحو أهمية إنشاء معاجم دلالية للغات الأخرى ومن هنا بدأت مشاريع ووردنت للغات الأخرى لينشأ العديد من المعاجم الدلالية المربوطة بووردنت [5] (باستخدام فهرس ووردنت لفهرسة السنستات الخاصة بها).

بالإضافة إلى الدور المهم في المعالجة اللغوية فإن لترجمة برينستون ووردنت أهمية بالغة وهي مع أننا نترجم ووردنت للغة العربية وبالتالي الحصول على قاموس دلالي إنجليزي - عربي يقابله قاموس دلالي عربي - إنجليزي فإننا بهذه الترجمة نحصل على عشرات القواميس الدلالية الأخرى أي أن ترجمة برينستون ووردنت للغة العربية يعني ترجمة الووردنت العربية لكل المعاجم الدلالية المترجمة الأخرى. هذا يمكن توضيحه كالتالي: إذا كانت سنست (أ) في ووردنت العربية مثلا هي ترجمة السنست (ب) في برينستون ووردنت وكانت (ج) هي ترجمة السنست (ب) بلغة أخرى كالألمانية، بالتالي فإن السنست (أ) هي ترجمة السنست (ج) من اللغة العربية إلى اللغة الألمانية والسنست (ج) هي ترجمة السنست (أ) من الألمانية إلى اللغة العربية.

بريستون ووردنت معجم دلالي خاص باللغة الإنجليزية ومستوحى من الثقافة الأمريكية. هذا يعني أن ترجمة هذا المعجم الدلالي للغات الأخرى يجب أن يراعي خصوصية اللغة والثقافة المراد الترجمة لها [6]. وبما أن برينستون ووردنت للغة واحدة فلم يلتفت المطورون إلى مشكلة الفجوة الدلالية [7, 8] بين اللغات المختلفة كذلك لا يوجد ضمان أن برينستون ووردنت خالية من الأخطاء وصحيحة مئة بالمئة. قد يكون هناك أخطاء بشرية حصلت أثناء تطوير المعجم أو أن تطوير المعجم أهمل بعض القواعد المهمة أو ربما لم ينتبه مطوورو المعجم لهذه القواعد [9].

نقوم في في هذا البحث بعرض أهم المشاكل التي تواجهنا عند تعريب برينستون ووردنت، ونشرح بعض المبادئ التي يجب مراعاتها خلال عملية التعريب ونعرف ما هي الجودة ونضع المعايير لها و نقدم الإطار العام الذي يضمن لنا تلك الجودة.

هذا البحث مقسم كالتالي. نعطي في القسم الأول لمحة عن البنية التنظيمية لووردنت. نسرد في القسم الثاني مشاريع تعريب برينستون ووردنت السابقة والحالية. في القسم الثالث نشرح أهم المشاكل عند تعريب برينستون ووردنت. في القسم الرابع نعرف الجودة ونضع المعايير الخاصة بها. في القسم الخامس نقدم الإطار العام للتعريب ونشرح المنهج العام المقترح لتعريب ووردنت والوصول لجودة عالية في عملية التعريب. في القسم السادس نطرح بعض النتائج التي وصلنا لها ونناقشها. في القسم السابع نضع خاتمة هذا البحث.

## 2   البنية التنظيمية لووردنت

الوحدة الأساسية في ووردنت هي السنست وهي مجموعة من المترادفات تشترك في معنى واحد. وتمثل كل سنست مفهوما محددا أو معنى معينا يتم توضيحه من خلال تعريف يوضح المعنى الذي تمثله هذه السنست. بالاضافة للتعريف قد تحتوي السنست على أمثلة توضيحية لكيفية استخدام المترادفات لتعطي معنى أو تعبر عن مفهوم السنست المقصود. المثال التالي هو السنست التي تعبر عن مفهوم سيارة:

**المترادفات**: {سيارة، مركبة}.

**التعريف**: (عربة آلية ذات أربع عجلات تستخدم في نقل الناس أو البضائع).

**أمثلة توضيحية**:  "أحتاج لسيارة للوصول للمنزل."  "شاهدت أحمد يركب مركبة بيضاء."

تعتمد البنية التنظيمية لووردنت على تقسيم مفردات ووردنت إلى خمسة فئات هي الأسماء والأفعال والصفات والأحوال والكلمات الوظيفية وتستخدم ووردنت في كل فئة عددا من العلاقات اللغوية لتنظيم العلاقات اللغوية وعدة علاقات دلالية لتنظيم العلاقات بين سنسنتات ووردنت نسرد فقط العلاقات التي تهمنا في تعريب ووردنت.

**علاقة الترادف**

تكون كلمتين أو أكثر مترادفات إذا كانت تلك الكلمات تعبر عن نفس المفهوم وتنتمي لنفس السنست. ينقسم الترادف إلى نوعين هما الترادف الحقيقي والترادف السياقي [2]. الترادف الحقيقي يحصل بين كلمتين اذا استطعنا استبدال هاتين الكلمتين في أي جملة دون تغير المعنى لتلك الجملة وهذا النوع نادر الوجود أما النوع الآخر وهو الذي تستخدمه ووردنت والمعاجم الأخرى فهو الترادف السياقي وهو اعتبار كلمتين مترادفتين حال وجود سياق واحد على الأقل يمكن استبدال الكلمتين فيه دون أن يتغير المعنى. على سبيل المثال هناك سياقات تكون كلمة مائدة مرادفة لكلمة طاولة مثل (وضع الطعام على الطاولة) أو (وضع الطعام على المائدة) ولكن هذا لا يتحقق في كل السياقات مثل أننا لا نستطيع استبدال كلمة طاولة بمائدة عند قولنا (يبدوا أن الاتفاق تم من تحت الطاولة).

## تعدد المعاني

تكون كلمة متعددة المعاني إذا كانت تعبر عن أكثر مفهوم مثل كلمة عين التي لها عدة معاني مثل (عضو البصر)، (عين ماء)، (جاسوس) أو غيرها من المعاني. بالتالي فإن كلمة عين تنتمي لأكثر من سنست. تحتوي ووردنت على الأنواع التالية لتعدد المعاني [12]:

- **المجانسة اللفظية**: وهو عندما تشترك الكلمات باللفظ وتختلفان بالمعنى مثل كلمة جمل التي تعني حيوان الجمل أو الجمل بمعنى الحبل الغليظ.

- **الاستعارة**: وهو التشبيه البليغ الذي يحذف أحد طرفيه مثل قولنا هو ثعلب لتشبيه شخصما بالمراوغة التي هي من صفات الثعلب.

- **الكناية**: لفظ لا يقصد منه المعنى الحقيقي وإنما معنى ملازما للمعنى الحقيقي مثل عندما يقول شخص (اتصلت بالمدرسة) فهو يقصد أنه اتصل بإدارة المدرسة أو (تأسست المدرسة قبل عشرين عام) فهو يقصد المدرسة كمؤسسة وليس مبنى المدرسة بينما (ذهبت إلى المدرسة) فالمقصود هنا المدرسة كبناية.

بالإضافة إلى هذه الأنواع تحتوي ووردنت على أنواع أخرى تعدد معاني تشير الكثير من الدراسات مثل[13] [14] [15] [16] وغيرها كما هوالحال في هذه الدراسة على عدم ملاءمة وجود هذه الأنواع في المعاجم وذلك لأنها تشكل صعوبة بالغة لخوارزميات التطبيقات اللغوية. السبب الرئيسي لوجود هذه الأنواع من تعدد المعاني الغير ملائمة هو تعدد المعاني التخصصي [17] أو استخدام جزء من اسم مركب [18] كمرادف لذلك الاسم مثل أن تكون كلمتا مركز و مركز تجاري مترادفتان في السنست التالية: {مركز تجاري، مركز، مركز تسوق} (مؤسسة تجارية تتكون من مجمع من المتاجر ذات المناظر الطبيعية التي تمثل التجار الرائدين) "لقد أمضوا العطلة في مراكز التسوق."

## 3    مشاريع تعريب ووردنت

بشكل عام تعتمد مشاريع ترجمة بريستون ووردنت على أحد النموذجين التاليين [19]:

- **نموذج الدمج:** يتم جمع السنستات في هذا النموذج من مصادر لغوية مثل المعاجم أولا وبعد ذلك يتم ربط هذه السنستات بالسنستات المقابلة لها بووردنت. على سبيل المثال استخدم هذا النموذج في مشروع الأنطولوجيا العربية [20،21،22] والووردنت الهندي [23].

- **نموذج التمدد:** تترجم السنستات في ووردنت مباشرة إلى اللغة المقصودة في هذا النموذج وهو النموذج المتبع في إنشاء معظم المعاجم الدلالية المربوطة بووردنت مثل الووردنت البولندي [24] وغيرها من المعاجم الدلالية.

يمتاز نموذج التمدد بسهولة استخدامه مقارنة بنموذج الدمج الذي يتطلب جهدا اضافيا ببناء السنستس قبل ربطها بووردنت. بنفس الوقت يتفوق نموذج الدمج على نموذج التمدد بقدرته على معالجة الاختلافات اللغوية والثقافية بين اللغة العربية والإنجليزية.

كانت أول محاولة لبناء شبكة كلمات عربية [25] قدمت نهجا آليا يعرب باسم SALAAM: SenseAssignment Leveraging Annotations And Multilinguality. ترجم في هذا المشروع 447 سنست للغة العربية. لا تحتوي النستات المترجمة في على المشروع على تعريفات أو أمثلة توضيحية.

بدأ في عام 2006 مشروع ثاني [26] لتعريب ووردنت حيث طور الباحثون ووردنت العربية AWN V1 باستخدام نموذج التمدد بترجمة 9618 سنست للغة العربية. تلى ذلك المشروع مشروع آخر [27] قام فيه الباحثون بإضافة ما يقارب 1700 سنست جديدة وتضمينها في نسخة جديدة لووردنت العربية AWN V2.

في عام 2012 بدأ مشروع معجم دلالي جديد هو مشروع الأنطولوجيا العربية. يرتكز المشروع على إنشاء هيكل تصنيفي (ontology) يمثل المفاهيم والكيانات الخاصة باللغة العربية ويتميز عن سابقيه باضافته تعريفات لغوية للسنتسات العربية ودقته في تعريف المفاهيم بمنهجية واضحة مما أدى إلى أن يكون هذا المشروع مرجعا جديدا ومتطورا للتطبيقات اللغوية والباحثين والمهتمين بمعالجة اللغة العربية.

هذا المشروع يختلف عن المشاريع السابقة بأنه ليس مشروع تعريب وإنما إنشاء أونطولوجيا مستقلة للغة العربية وهو يختلف عن ووردنت بتركيزه على تقديم أونطولوجيا للغة العربية تعتمد على التعريفات الفلسفية والانضباط المنطقي للمصطلحات مذلك تركيزه على انضباط التعريفات اللغوية وصياغتها بشكل يضمن سهولة تتبع دلالاتها وصفاتها بشكل منطقي. بالاضافة إلى ذلك فالأنطولوجيا العربية تستخدم هيكلا تنظيميا مختلفا عن ووردنت يساعد في حل مشاكل الغموض الدلالي التي تواجه الاستخدامات الدلالية للأنطولوجيا في المشاريع الإلكترونية مثل تطبيقات الحكومات الإلكترونية. بالطبع يمكن ربط الأونطولوجيا العربية أو أجزاء منها بووردنت من خلال نموذج الدمج.

بالاضافة لتلك المشاريع ما زالت الجهود مستمرة لتطوير وتحسين المعاجم الدلالية العربية حيث تصدر أبحاث وجهود جديدة كل عام تقريبا كان آخرها مشروع يهدف إلى استخدام الترجمة الآلية العميقة لترجمة برينستون ووردنت إلى العربية [28]. مع أن هذه الفكرة قد تقدم حلا مؤقتا لتعريب الووردنت وهي مشكلة خلو الساحة للغة العربية إلا أن الترجمة الآلية مهما كانت دقتها فإنها لن تكون ملائمة كمرجع لغوي دون تدقيق بشري خصوصا أن الترجمة الآلية لن تستطيع تقديم قاموسا يعالج الاختلافات اللغوية والثقافية.

بالنسبة لمشاريع SALAAM ووررردنت العربية بنسختيها فإننا نواجه بالإضافة إلى قلة عدد السنستات المترجمة مشكلة خلو الترجمة من التعريفات والأمثلة التوضيحية واكتفاء المطروحين بترجمة المترادفات. لهذا السبب فإنه من الصعب جدا الحكم على صحة ترجمة المترادفات وكذلك فإن الفائدة من هذه الترجمة تقل بغياب ترجمة هذه العناصر المهمة. بالإضافة إلى ذلك لا تعالج تلك المشاريع الفجوة الدلالية أو تتطرق لذكرها.

## 4  مشاكل تعريب ووردنت

نطرح في الأقسام التالية أهم المشاكل التي يجب مراعاتها في عملية التعريب للوصول إلى جودة عالية.

### 4.1  جودة الترجمة

كل المشاريع المتعلقة بتعريب ووردنت لم تعط أي معايير كاملة للحكم على جودة الترجمة ففي حين أن مشروع الأنطولوجيا العربية الذي هو ليس مشروع تعريب كما أسلفنا سابقا أعطى معايرا لكيفية كتابة تعريف السنست إلا أننا لم نجد تعريفا شاملا أو معايير كاملة في مشاريع التعريب السابقة تعرف لنا جودة الترجمة.

في بحث سابق [29] عرفنا جودة الترجمة من خلال معيارين أساسيين وهما صحة ترجمة واكتمال ترجمة السنست. نعني بصحة الترجمة التالي:

**صحة ترجمة المترادفات:** أن تكون كل المترادفات المترجمة صحيحة وتعبر عن معنى السنست. هذا يعني خلو مجموعة المترادفات من الأخطاء.

**صحة ترجمة تعريف السنست:** تعريف السنست صحيح وواضح و يعبر عنها.

**صحة ترجمة الأمثلة التوضيحية:** أن تكون الأمثلة التوضيحية صحيحة تعبر عن معنى السنست وكذلك أن تكون تلك الأمثلة مناسبة للسياق الثقافي واللغوي للغة المراد الترجمة لها.

ونعرف اكتمال الترجمة كالتالي:

**اكتمال ترجمة المترادفات:** أن تحتوي السنست على جميع المترادفات التي تعبر عن مفهوم السنست.

**اكتمال ترجمة التعريف:** أن يحتوي التعريف على العناصر الأساسية لفهم المعنى المقصود من السنست و يفهم من خلاله الفرق بين السنست و السنستات القريبة بالمعنى منها.

**اكتمال ترجمة الأمثلة التوضيحية:** أن يكون لكل من مترادفات السنست مثالا توضيحيا لاستخدام تلك المفردة بذلك المعنى.

## 4.2   الفجوة المعجمية

إن أهم مشكلة تواجهنا عند ترجمة المعاجم هي مشكلة الاختلافات الثقافية واللغوية والتي يطلق عليها أحيانا اسم التنوع اللغوي [27]. على سبيل المثال فإن الكلمة الإنجليزية، cousin والتي تعني ابن العم أو إبن الخال، ليس لها أي كلمة مماثلة في اللغة العربية. في المقابل، فإن الكلمة العربية عم ، والتي تعني شقيق الوالد، لا وجود لها في اللغة الإنجليزية. أمثلة أخرى نجدها في الألوان أو الأفعال. على سبيل المثال to bike في برينستون ووردنت التي تعني "ركب الدراجة" لا يوجد ما يقابله في كل اللغات مثل مثل اللغة العربية أو الإيطالية.

يشير اللغويون إلى مثل هذه الحالات بالفجوة المعجمية وهي عند غياب مفهوم معجمي في لغة ووجوده في لغة أخرى. عادة ما يلجأ المترجمون في مثل هذه الحالات إلى استخدام معنى مماثل باستخدام أشباه الجمل [30].

بالنسبة للغة العربية فإن ظاهرة الفجوة المعجمية نجدها بشكل كبير عند ترجمة الأحوال. على سبيل المثال كلمة injuriously أو adjectively لا نجد لها حالا مقابلا باللغة العربية. بالاضافة للفجوة المعجمية نجد الكثير من الحالات التي يشيع استخدام أشباه الجمل عند التعبير عنها. على سبيل المثال فإن الشائع هو استخدام كل عام أو كل شهر أو كل أسبوع للتعبير عن سنو يا أو شهر يا أو أسبوعيا. مثال آخر هو استخدام بشكل (الصفة المشتق منها الحال) للتعبير عن الحال مثل بشكل جنوني أو بشكل مثير.

عدم ادراج أشباه الجمل هذه مع شيوع استخدامها خلال عملة التعريب قد يؤدي إلى خلل في جودة التعريب وكذلك إلى التحيز نحو اللغة الإنجليزية وهذا سيقلل أيضا من جودة التطبيقات التي تعتمد على التعريب، مثل أدوات الترجمة.

## 4.3   تعدد المعاني

تعدد المعاني مشكلة معروفة في برينستون ووردنت وقد تم بحث هذه المشكلة في العديد من الدراسات كما أشرنا سابقا. نشكلة تعدد المعاني التي تواجهنا خلال ترجمة ووردنت ليست التجانس، أو المجاز، أو الكناية التي هي ضرورية في المعاجم ولكن تعدد المعاني التخصصي وتعدد معاني الأسماء المركبة حيث يسبب هذان النوعين صعوبة في استخدام ووردنت مصدر لمعالجة اللغة الطبيعية. كمثال على تعدد معاني الأسماء المركبة، تحتوي كلمة head على أكثر من 30 سنست في ووردنت. مثال آخر على تعدد معاني الأسماء المركبة هو كلمة، center والتي تحتوي على 18 سنست. بالنسبة لتعدد المعاني التخصصي فإن كلمة turtledove هي كلمة متعددة المعاني لأنها تنتمي إلى مجموعتي المرادفات التاليتين:

(1) Australian turtledove, Stictopelia cuneata, turtledove small Australian dove

(2) turtledove any of several Old World wild doves

بالطبع، من الممكن استخدام كلمة turtledove للإشارة إلى أي نوع من أنواع turtledoves عندما يكون واضحا من السياق أي نوع من turtledoves نتحدث عنه. في الوقت نفسه، فإن إضافة كلمة turtledove كمرادف لجميع أنواع turtledoves في المورد المعجمي أمر غير مجدٍ ويجعل من الصعب استخدام المورد. وفقا لبحثنا [18] فإن السياق وحده كافي لإزالة الغموض عن معنى الكلمة لهذين النوعين من تعدد المعاني ولا يتطلب تضمين كل هذه المعاني المحتملة في مورد معجمي لأنها تؤدي إلى مشكلة تضاعف عدد المعاني مما يجعل من الصعب جدا استخدام مثل هذه الموارد في معالجة اللغة الطبيعية.

## 5    تعريب ووردنت

نقدم في هذا القسم الاطار العام لتعريب ووردنت مع شرح تفصيلي للخطوات اللازم اتخاذها للوصول لترجمة عالية الدقة.

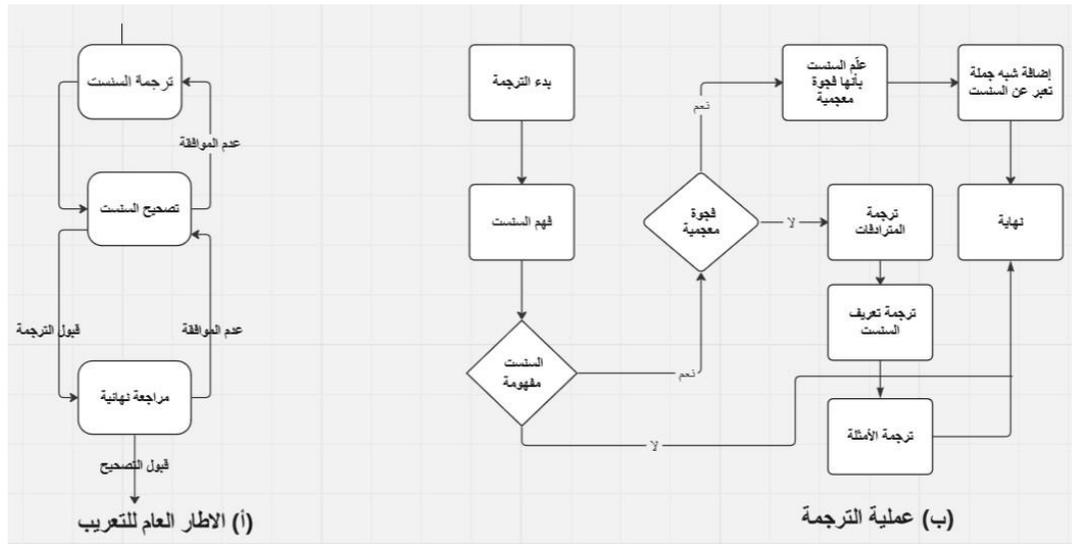

شكل 1: تعريب السنسنس

### 5.1    الاطار العام لعملية التعريب

لضمان الوصول إلى أعلى جودة للتعريب فإن العملية التعريب تنقسم إلى المراحل التالية كما هو موضح في الشكل 1 (أ):

**مرحلة الترجمة:** يقوم مترجم متخصص بترجمة السنست حسب المعايير والخطوات التي سنشرحها في القسم التالي.

**مرحلة التصحيح:** يقوم مترجم آخر بمراجعة صحة واكتمال السنست حسب نفس المعايير التي يستخدمها المترجم الأول.

النتيجة هي قبول الترجمة أي أن المصحح حكم على صحة واكتمال الترجمة. عند رفض الترجمة لعدم صحة أو اكتمال أجزاء من الترجمة فإن المصحح يضع ملاحظاته التي يطلب من المترجم الأول مراعاتها وتُرجع الترجمة للمترجم الأول حتى التوافق التام بينهما على صحة واكتمال الترجمة.

**مرحلة المراجعة النهائية:** في هذه المرحلة يقوم متخصص باللغة العربية بمراجعة الترجمة النهائية التي توافق عليها المترجم والمصحح. تعتبر الترجمة صحيحة ومكتملة في حال حكم المتخصص على ذلك. عند رفض الترجمة تعاد الترجمة مع ملاحظات المتخصص ليقوم المصحح بالتعديل عليها. تستمر هذه العملية حتى التوافق النهائي بين المتخصص والمراجع.

### 5.2 مرحلة الترجمة

يقوم بالمرحلة الأولى من التعريب مترجمان حاصلان على درجة البكالوريوس على الأقل في مجال الترجمة (الإنجليزية العربية). قبل الترجمة، ندرب المترجمين كما هو موضح في الشكل 1(ب) ونشرحه في الفقرات التالية:

**1 فهم معنى السنست:** الهدف من هذه المرحلة هو ضمان أن يكون للمترجم فهما واضحا للسنست التي يريد ترجمتها. قد يكون سبب عدم فهم السنست هو عدم فهم بعض مترادفاتها أو عدم فهم التعريف الإنجليزي للسنست باللغة اللغة. يمكن هنا للمترجم الاستعانة بالمصادر الإضافية التالية للتأكد من فهم السنست:

- استخدم الموارد الخارجية مثل القواميس وويكيبيديا.

- يسمح للمترجم بترك السنست دون ترجمة في حال أن المترجم لم يفهمها مع ترك ملاحظة أنه لم يفهم السنست.

**2 تحديد الفجوة المعجمية:** قد كون سبب الفجوة المعجمية أن لا يكون مفهوم السنست معروفا في الثقافة أو اللغة العربية أو عند عدم وجود كلمة في اللغة العربية تعبر عن مفهوم السنست. في هذه المرحلة يقرر المترجم إذا كان هناك كلمة أو كلمات تعبر عن مفهوم السنست في اللغة العربية أو أن السنست فجوة معجمية بناء على فهم المترجم للسنست كما تم شرحه في الخطوة السابقة. ينفذ المترجم التعليمات في (أ) إذا كانت السنست فجوة معجمية، عدا ذلك فإنه يقوم بالخطوات في (ب) و (ج).

**الخطوة أ - معالجة الفجوة المعجمية:** في هذه الخطوة، يُطلب من المترجم وضع علامة أن السنست فجوة معجمية ووضع شبه جملة باللغة العربية تعبر عن مفهوم السنست. على سبيل المثال تعتبر السنست التالية فجوة معجمية يقابلها شبه الجملة "بشكل معبر":

expressively with expression; in an expressive manner "she gave the order to the waiter, using her hands very expressively"

**الخطوة ب - ترجمة السنست:** بعد أن يتأكد المترجم أن السنست ليست فجوة معجمية، يقوم بترجمة السنست حسب الخطوات التالية:

**i ترجمة تعريف السنست:** يحرص المترجم على جودة الترجمة وأن تكون الجودة بنفس جودة التعريف باللغة الإنجليزية على الأقل. يجب أن يبتعد المترجم عن غموض بالتعريف وأن التعريف يعبر عن معنى السنست بشكل واضح.

**ii ترجمة المترادفات:** هنا يتنبه المترجم إلى أن عدد المترادفات العربية قد تكون أقل أو أكثر من مثيلاتها باللغة الإنجليزية. كذلك ينتبه المترجم إلى أن المترادفات الإنجليزية قد تكون غير مكتملة مما يعني أنها قد لا تحتوي على جميع المترادفات التي تعبر عن مفهوم السنست.

لترجمة مرادفات مجموعة المرادفات، نمر بالمراحل التالية:

- **ترجمة المرادفات الإنجليزية:** نتيجة هذه الخطوة هي عدد من المرادفات العربية المقابلة للمرادفات الإنجليزية.

- **إضافة مرادفات عربية:** يجمع المترجم لكل كلمة عربية من الخطوة السابقة كل مرادفات تلك الكلمة.

- **التحقق من صحة المرادفات العربية:** يقوم المترجم باستبعاد الكلمات التي لا تعبر عن معنى السنست اعتمادا على تعريفها والقدرة على إيجاد مثال لمل مفردة يعبر عن مفهوم السنست.

- **جمع المترادفات:** تعتبر كل كلمة ناتجة من المرحلة السابقة من مترادفات السنست إذا كان يقابلها مثال توضيحي واحد على الأقل. نقوم في هذا المرحلة أيضا باستبعاد أي كلمة ستؤدي إلى تعدد معاني تخصصي أو تعدد معاني الاسم المركب. على سبيل المثال، تم استبعاد "جسم" من {جسم، جسم طبيعي} المقابلة للسنست:

object, physical object a tangible and visible entity; an entity that can cast a shadow "it was full of rackets, balls and other objects"

- **ترتيب المصطلحات العربية:** نرتب هنا المترادفات حسب شيوعها في التعبير عن معنى السنست باستخدام الأمثلة التوضيحية من حيث سهولة وسلاسة استخدام المترادفات في الأمثلة التوضيحية.

**3. ترجمة الأمثلة التوضيحية:** تساهم الأمثلة التوضيحية للمترادفات في فهم أوضح لكيفية استخدام السنست. نستخدم نفس الأمثلة التي تم صياغتها أثناء مرحلة ترجمة المترادفات كأمثلة للمترادفات. يجدر الإشارة إلى أننا لا نترجم فقط الأمثلة الموجودة في المترادفات الإنجليزية بل نشترط أن يكون لكل كلمة من مترادفات السنست مثال توضيحي على الأقل حتى لو لم تحتوي المترادفات الإنجليزية على أمثلة على الإطلاق. ترتيب الأمثلة التوضيحية للمترادفات يتم حسب ترتيب المترادفات في الخطوة السابقة.

### 5.3 مرحلتا التصحيح والمراجعة النهائية

في مرحلتي التصحيح والمراجعة النهائية نستخدم نفس الخطوات لقبول الترجمة أو رفضها وفيما يلي نشرح الخطوات في كلتا المرحلتين:

**1 التحقق من صحة الفجوات المعجمية:** يتحقق المصحح من صحة الفجوات المعجمية التي حددها المترجم في المرحلة الأولى. نتيجة هذه الخطوة إما الموافقة على أن السنست الإنجليزية يقابلها فجوة معجمية في اللغة العربية أو رفض الفجوة المعجمية من خلال تقديم مترادفات عربية تعبر عن مفهوم السنست وبالتالي ترجمة كاملة للسنست في اللغة العربية.

**2 التحقق من صحة تعريف السنست:** يتحقق المصحح هنا من أن تعريف السنست العربي يعبر عن المعنى المقصود للسنست في اللغة الإنجليزية. إذا كان الجواب نعم يتحقق بعدها من سلاسة وسهولة التعريف بالإضافة إلى التحقق من الأخطاء اللغوية والإملائية.

**3 التحقق من صحة واكتمال المترادفات:** يتحقق المصحح من أن المترادفات صحيحة و لا تحتوي على مترادفات خاطئة (لا تعبر عن مفهوم السنست) وكاملة أي لا يوجد مترادفات ناقصة. يستخدم المصحح الأمثلة التوضيحية للتحقق من صحة المترادفات.

**4 التحقق من صحة الأمثلة التوضيحية:** يتحقق المصحح من أن كل كلمة من كلمات المترادفات يقابلها مثال واحد على الأقل. كذلك يتحقق المصحح من أن الأمثلة طبيعية غير متكلفة وتعبر عن المعنى المقصود للسنست.

في حالة وجود أخطاء أو اختلافات في تقييم الترجمة، تعاد السنستات إلى المترجم مع تعليقات المصحح. السنستات المقبولة ترسل للخبير اللغوي للمراجعة النهائية.

في مرحلة المراجعة النهائية، تسند عملية المراجعة لخبير لغوي عربي يقوم بنفس الخطوات التي قام بها المصحح الأول. الفرق بين البيانات التي ترسل للخبير اللغوي والمصحح هي أن الخبير اللغوي لا يطلع على الأصل الإنجليزي وإنما على السنستات باللغة العربية فقط في حال عدم وجود فجوة معجمية. بالنسبة للفجوات المعجمية فإن الخبير يطلع على السنست باللغة الإنجليزية أيضا للتأكد من صحة وجود فجوة معجمية في اللغة العربية. في حالة وجود أخطاء أو عدم توافق مع التصحيح تعاد السنستات الغير مقبولة للمصحح مع ملاحظات الخبير اللغوي.

| قسم الكلام | عدد السنستات | عدد المترادفات |
|---|---|---|
| الأسماء | 6516 | 13659 |
| الأفعال | 2507 | 5878 |
| الصفات | 446 | 761 |
| الأحوال | 107 | 262 |
| العدد الكلي | 9576 | 20560 |

جدول 1: عدد السنستات والكلمات الموجودة في الووردنت العربية AWN1 مقسمة حسب أقسام الكلام

## 6   نتائج ونقاش

نهدف من عملية التعريب المقترحة تعريب أكبر قدر ممكن من بريستون ووردنت مقسمة لمراحل. في المرحلة الأولى [29] قمنا بتجربة هذه العملية على تصحيح وتطوير ووردنت العربية AWN V1[1]. فيما يلي بعض الإحصائيات الخاصة بهذه المرحلة.

| | أسماء | أفعال | صفات | أحوال | عدد كلي |
|---|---|---|---|---|---|
| مترادفات جديدة | 2581 | 64 | 72 | 9 | 2726 |
| مترادفات تم استثناؤها | 6050 | 2387 | 223 | 91 | 8751 |
| تعاريف تم اضافتها | 6511 | 2258 | 446 | 107 | 9322 |
| أمثلة تم اضافتها | 7597 | 3620 | 782 | 205 | 12204 |
| فجوات معجمية تم تحديدها | 28 | 187 | 0 | 21 | 236 |
| أشباه جمل تم اضافتها | 364 | 275 | 0 | 62 | 701 |

جدول 2: إحصائيات الإضافة والحذف في AWN1

قمنا بتجريب طريقة التعريب المقترحة على 6516 اسما و2507 فعلا و446 صفة و107 حالا (انظر الجدول 1 لمزيد من التفاصيل). كانت النتيجة هي إضافة 2726 مرادفا جديدا و9322 تعريفا جديدا، و12204 مثالا توضيحيا جديدا. كما تم تحديد 236 فجوة معجمية وإدراج 701 شبه جملة. كان عدد المترادفات التي تم استثناؤها 8751 كلمة كما هو موضح في الجدول 2. من خلال هذه النتائج يتبين لنا أهمية طريقة التعريب المقترحة خصوصا إضافة تعريفات السنست والأمثلة التوضيحية للمترادفات كآلية لإضافة مترادف جديد أو استثناء مترادف غير ملائم ولا يعبر عن معنى السنست.

## 7   خاتمة

لا بد لنا عند تعريب المعاجم الدلالية من وضع نهج واضح للتعريب يضمن لنا الوصول إلى أعلى جودة. بالاضافة إلى هذا المنهج يتوجب علينا أيضا أن نكون ملمين بالمشاكل الموجودة بالمعجم المراد تعريبه وأيضا الإلمام بالحلول لتلك المشاكل. اقترحنا في هذا البحث منهجا شاملا لتعريب المعجم الدلالي بريستون ووردنت وعند تجربة هذا المنهج تبين لنا أن ترجمة المعجم لا يعني بحال الاكتفاء بترجمة المترادفات وإهمال ترجمة التعريفات والأمثلة التوضيحية لأنه بدون تعريب تلك العناصر فإننا نفتقد إلى معايير للحكم على جودة الترجمة وبنفس الوقت لا يوجد أي ضمان آخر يثبت لنا صحة واكتمال الترجمة من عدمها.

نهدف إلى مواصلة العمل على تطوير منهج الترجمة المقترح في هذا البحث واستخدامه في إكمال تعريب ووردنت حيث أننا قاربنا حاليا على الانتهاء من ترجمة وتصحيح جميع الأحوال الموجودة في ووردنت وعددها 3642 حال. بعد انتهاء التصحيح والمراجعة النهائية سننتقل إلى الصفات والأفعال والأسماء لنترجم سنستات جديدة لم تترجم من قبل. نأمل أن ننتهي من تعريب

---

[1] https://github.com/HadiPTUK/AWN3.0

جميع الأحوال والصفات والأفعال في ووردنت خلال عام من الآن واتاحتها كمصدر مفتوح للمجتمع العلمي للاستفادة منها وكذلك تعديل هذه النتائج وتحسينها. في المرحلة الأخيرة سنقوم بتعريب الأسماء التي تعد القسم الأكبر في برينستون ووردنت.

# المراجع

[1] Miller, G. A. Wordnet: a lexical database for english. Communications of the ACM 38, 11 (1995), 39–41.

[2] Miller, G. A. - 10-nouns in wordnet : A lexical inheritance system.

[3] Miller, G. A., Beckwith, R., Fellbaum, C., Gross, D., and Miller, K. J. Introduction to wordnet: An online lexical database. International journal of lexicography 3, 4 (1990), 235–244.

[4] Morato, J., Marzal, M., Llorens, J., and Moreiro, J. Wordnet applications. Proceedings of the 2nd Global Wordnet Conference 2004 (04 2004).

[5] Vossen, P. Eurowordnet, 1999, PJTM.

[6] Freihat, A. A., Khalilia, H. M., Bella, G., and Giunchiglia, F. Advancing the Arabic WordNet: Elevating content quality. In Proceedings of the 6th Workshop on Open-Source Arabic Corpora and Processing Tools (OSACT) with Shared Tasks on Arabic LLMs Hallucination and Dialect to MSA Machine Translation @ LREC-COLING 2024 (Torino, Italia, May 2024), H. Al-Khalifa, K. Darwish, H. Mubarak, M. Ali, and T. Elsayed, Eds., ELRA and ICCL, pp. 74–83.

[7] Giunchiglia, F., Batsuren, K., and Freihat, A. A. One world seven thousand languages. In Proceedings 19th International Conference on Computational Linguistics and Intelligent Text Processing, CiCling2018, 18-24 March 2018 (2018).

[8] Schuler, K. K. VerbNet: A broad-coverage, comprehensive verb lexicon. University of Pennsylvania, 2005.

[9] Bentivogli, L., and Pianta, E. Looking for lexical gaps. In Proceedings of the ninth EURALEX International Congress (2000), Stuttgart: Universität Stuttgart, pp. 8–12.

[10] Lehrer, A. Notes on lexical gaps. Journal of linguistics 6, 2 (1970), 257–261.

[11] Khalilia, H., Freihat, A. A., and Giunchiglia, F. The quality of lexical semantic resources: A survey. In Proceedings of the 4th International Conference on Natural Language and Speech Processing (ICNLSP 2021) (2021), pp. 117–129.

[12] Freihat, A. A., Giunchiglia, F., and Dutta, B. A taxonomic classification of wordnet polysemy types. In Proceedings of the 8th Global WordNet Conference (GWC) (2016), pp. 106–114.

[13] Buitelaar, P. P. CoreLex: systematic polysemy and underspecification. Brandeis University, 1998.

[14] Navigli, R. Word sense disambiguation: A survey. ACM Comput. Surv. 41 (02 2009).

[15] Freihat, A. A. An organizational approach to the polysemy problem in wordnet. PhD thesis, University of Trento, 2014.

[16] Snow, R., Prakash, S., Jurafsky, D., and Ng, A. Learning to merge word senses. pp. 1005–1014.

[17] Freihat, A. A., Giunchiglia, F., and Dutta, B. Solving specialization polysemy in wordnet. Int. J. Comput. Linguistics Appl. 4, 1 (2013), 29–52.

[18] Freihat, A. A., Dutta, B., and Giunchiglia, F. Compound noun polysemy and sense enumeration in wordnet. In Proceedings of the 7th International Conference on Information, Process, and Knowledge Management (eKNOW) (2015), pp. 166–171.

[19] Khalilia, H., Freihat, A. A., Giunchiglia, F., et al. The dimensions of lexical semantic resource quality. In Proceedings of the Second International Workshop on NLP Solutions for Under Resourced Languages (NSURL 2021) co-located with ICNLSP 2021 (2021), ACL Anthology, pp. 15–21.

[20] Jarrar, M. Position paper: towards the notion of gloss, and the adoption of linguistic resources in formal ontology engineering. In Proceedings of the 15th international conference on World Wide Web (2006), pp. 497–503.

[21] Jarrar, M., and Amayreh, H. An arabic-multilingual database with a lexicographic search engine. In Natural Language Processing and Information Systems: 24th International Conference on Applications of Natural Language to Information Systems, NLDB 2019, Salford, UK, June 26–28, 2019, Proceedings (Berlin, Heidelberg, 2019), Springer-Verlag, p. 234–246.

[22] Jarrar, M. The arabic ontology–an arabic wordnet with ontologically clean content. Applied ontology 16, 1 (2021), 1–26.

[23] Bhattacharyya, P. IndoWordNet. In Proceedings of the Seventh International Conference on Language Resources and Evaluation (LREC'10) (Valletta, Malta, May 2010), European Language Resources Association (ELRA).

[24] Piasecki, M., Broda, B., and Szpakowicz, S. A wordnet from the ground up. Oficyna Wydawnicza Politechniki Wrocławskiej Wrocław, 2009.

[25] Diab, M. The feasibility of bootstrapping an arabic wordnet leveraging parallel corpora and an english wordnet. In Proceedings of the Arabic Language Technologies and Resources, NEMLAR, Cairo (2004).

[26] Elkateb, S., Black, W., Rodríguez, H., Alkhalifa, M., Vossen, P., Pease, A., and Fellbaum, C. Building a WordNet for Arabic. In Proceedings of the Fifth International Conference on Language Resources and Evaluation (LREC'06) (2006), European Language Resources Association, pp. 29–34.

[27] Regragui, Y., Abouenour, L., Krieche, F., Bouzoubaa, K., and Rosso, P. Arabic wordnet: New content and new applications. In Proceedings of the 8th Global WordNet Conference (GWC) (2016), pp. 333–341.

[28] Souci, M. D. E., Cherifi, Y., Berkani, L., Ameur, M. S. H., and Guessoum, A. Enrichment of Arabic WordNet using machine translation and transformers. In Proceedings of the 6th International Conference on Natural Language and Speech Processing (ICNLSP 2023) (Online, Dec. 2023), M. Abbas and A. A. Freihat, Eds., Association for Computational Linguistics, pp. 333–340.

[29] Khalilia, H., Bella, G., Freihat, A. A., Darma, S., and Giunchiglia, F. Lexical diversity in kinship across languages and dialects. Frontiers in Psychology 14 (2023).

[30] Bentivogli, L., and Pianta, E. Beyond lexical units: Enriching WordNets with phrasets. In 10th Conference of the European Chapter of the Association for Computational Linguistics (Budapest, Hungary, Apr. 2003), A. Copestake and J. Hajič, Eds., Association for Computational Linguistics.

[31] Anthony McEnery and others. The emille/ciil corpus, 2004.

[32] Batsuren, K., Ganbold, A., Chagnaa, A., and Giunchiglia, F. Building the mongolian wordnet. In Proceedings of the 10th global WordNet conference (2019), pp. 238–244.

[33] Beckwith, R., Fellbaum, C., Gross, D., and Miller, G. A. Wordnet: A lexical database organized on psycholinguistic principles. In Lexical Acquisition. Psychology Press, 2021, pp. 211–232.

[34] Farreres, J., Rodríguez, H., and Gibert, K. Semiautomatic creation of taxonomies. In COLING-02: SEMANET: Building and Using Semantic Networks (2002).

[35] Khalid Choukri and Niklas Paullson. The orientel moroccan mca (modern colloquial arabic) database, 2004.

[36] Speecon Consortium. Catalan speecon database, 2011.

[37] Tufis, D., Cristea, D., and Stamou, S. Balkanet: Aims, methods, results and perspectives. a general overview. Romanian Journal of Information science and technology 7, 1-2 (2004), 9–43.